\documentclass{article}
\usepackage{spconf,amsmath,epsfig}
\usepackage{amssymb}
\usepackage{bm}
\usepackage{mathtools}
\usepackage{float}
\usepackage{graphicx}
\usepackage{jabbrv}

\let\OLDthebibliography\thebibliography
\renewcommand\thebibliography[1]{
  \OLDthebibliography{#1}
  \setlength{\parskip}{0pt}
  \setlength{\itemsep}{0pt plus 0.3ex}
}

\renewcommand\paragraph[1]{{\boldmath\bfseries{#1}~:}}

\begin{document}\sloppy

\def\x{{\mathbf x}}
\def\L{{\cal L}}

\title{Modular Blended Attention Network for Video Question Answering}
%
\name{Mingjie Zhou}
\address{}

\maketitle

\begin{abstract}
  In multimodal machine learning tasks, it is due to the complexity of the assignments that the network structure, in most cases, is assembled in a sophisticated way. The holistic architecture can be separated into several logical parts according to the respective ends that the modules are devised to achieve. As the number of modalities of information representation increases, constructing ad hoc subnetworks for processing the data from divergent modalities while mediating the fusion of different information types has become a cumbersome and expensive problem. In this paper, we present an approach to facilitate the question with a reusable and composable neural unit; by connecting the units in series or parallel, the arduous network constructing of multimodal machine learning tasks will be accomplished in a much straightforward way. Additionally, through parameter sharing (weights replication) among the units, the space complexity will be significantly reduced. We have conducted experiments on three commonly used datasets; our method achieves impressive performance compared to several video QA baselines.
\end{abstract}
\section{Introduction}
In current multimodal machine learning tasks, because of the diversity of data modalities, numerous heterogeneous subnetworks designed for processing the data of different modalities have been proposed; combined with holistic neural networks in an end-to-end training manner, excellent results have been achieved. Without considering the huge computational power and cumbersome network structure design procedures, this may be an attractive solution. Though it is the current mainstream method, with the increase of task complexity and data modalities, the structural design of neural networks has gradually become an extremely energy-consuming task.

On the other hand, with the development of machine learning, many more complex tasks have been given, and these tasks contain more abstract logic and higher-level inferences. Solving these problems requires the ability to deal with complex reasoning processes and analyze the subtle interference between massive amounts of data, which directly leads to an increase in the number of neural network parameters. It is because that, with the similar network structures, the parameter amount of the neural network has a positive correlation with the range of the function set that can be simulated; under the condition of sufficient training data, a larger function set range guarantees the feasibility of completing the task, but at the same time, it also greatly increases the complexity of the calculation. In recent years, most of the neural network models have increasingly higher requirements for the hardware environment, and many studies are trapped here and cannot be carried out. The optimization of the computational complexity of the neural network has become an urgent task.

One possible way of addressing the problem is developing reusable modular network. Ellefsen et al.~\cite{ellefsen2015neural} prove that modularity in natural animal brains enables them to learn new skills without forgetting old skills; it may be a promising approach to devise artificial neural networks that can learn a variety of different skills.
A modular system built according to the methodological guidance of dividing and conquering not only has less resource occupation; in design, the reusable modular structure also greatly reduces the complexity of the task.


In this paper, we propose a universal modular network. Our universal module completes the data fusion process between different modalities and the reasoning of chronological clues with several attention mechanisms. By standardizing the input and output format of the network module, the module is capable of being replicated and layered straightforwardly; and through the network parameter sharing between the modules with similar functionality, the parameter amount of the network is considerably reduced. We verify the effectiveness of our universal module on the mainstream video question answering datasets and our method achieves competitive results on several tasks.

\section{Related Work}
Visual-linguistic models are dedicated to gaining fused representations of vision and natural language features; it conventionally involves two or even more modalities.

Recently, a lot of works~\cite{zhang2021fusing,wang2021dualvgr,lei2021less} follow the appearance-motion paradigm to encode the visual information. These video-and-language works tend to first apply pre-trained 2D convolutional neural networks (CNNs) like ResNet~\cite{he2016deep}, to first exploit sketchy appearance information from frames and utilize 3D CNNs, such as ResNeXt~\cite{xie2017aggregated}, to handle the motion feature of a clip or leverage recurrent neural network (RNN) to obtain the dynamic information directly from the frame sequence. Likewise, the linguistic representations are embedded with the pre-trained encoder (e.g., GloVe~\cite{pennington2014glove}) and then these embeddings are passed into an RNN to get context-aware representations. Then, these primitive modality-specific features extracted with general-purpose pre-trained neural networks are fused into joint embeddings to conduct the final decision-making process.

According to the location of the phase where multimodal representations are fused in the network architecture, these works can be grouped as early fusion and late fusion. Late fusion refers to the methods~\cite{li2020hero,sun2020learning} that feed the coarse multimodal features into modality-specific subnetworks to get the fine-grained features separately and then integrate the outcomes with the late fusion modular. The problems are that these methods handle multimodal interactions at a summary level and therefore the correlations of the features inside each modality have been neglected. In addition, designing ad hoc subnetworks for each modality is a time-consuming job and not suitable to extend to the assignments with more available modalities. More recently, a large number of early fusion methods~\cite{lei2020tvqa+,wang2021dualvgr} have been proposed.

Finetuning pre-trained models on demanded datasets is becoming a dominant path to solving the video question answering task. In~\cite{seo2021look}, they achieve remarkable results by leveraging a co-attentional multimodal video transformer model pre-trained on HowTo100M. Yang et al.~\cite{yang2020just} leverage a question generation transformer and automatically generate the HowToVQA69M dataset for pre-training and significantly outperform the state-of-the-art methods on several benchmarks. Unfortunately, these methods demand massive computational resources.

\section{Modular Neural Network}
\subsection{Generic Neural Network Module}

\begin{figure}[th]
  \centering
  \includegraphics[width=0.95\linewidth]{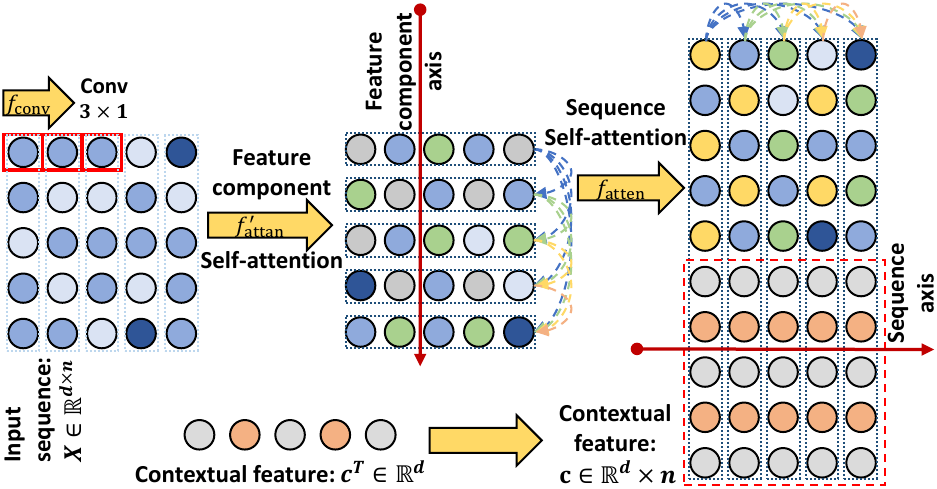}
  \caption{Our generic neural network module (GNNM) that consists of there parts: (1) chronological convolution, (2) feature component attention and (3) hybrid attention.}
  \label{fig:1}
\end{figure}

We have designed a generic neural network module (GNNM), as shown in Fig.~\ref{fig:1}, to handle the fusion of the feature embeddings that are projected from different modalities. The inputs of the module are a sequence of $n$ feature vectors, ${X}=(\bm{x_1}, \dotsc, \bm{x_n})\in\mathbb{R}^{d \times n}$, and a contextual vector, $\bm{c}\in\mathbb{R}^d$. As for the outputs, there are two optional forms available: (1) an array of vectors, ${Y}=(\bm{y_1},\dotsc, \bm{y_n})\in\mathbb{R}^{d \times n}$, that is of the shape of the input sequence ${X}$ or (2) a single vector, $\bm{y^\prime}\in\mathbb{R}^d$, that is aggregated from ${Y}$ according to the attention weights that are calculated prior to output phase. The module can be logically divided into three functions.

First, considering the similarity of chronologically adjacent features, we perform convolution-based operations on the feature map, ${X}\in\mathbb{R}^{n\times d}$, formed by the feature sequence. First, we use a kernel of $3\times 1$ to perform convolution operation on the feature map ${X}$; the result obtained is further processed by the softmax function along the time axis. Then, we used the softmax outcome as a mask to conduct the Hadamard product with the original feature map. Because the height of the convolution kernel is one, this operation does not induce the fusion of different feature attributes, but only attempts to capture some vital pieces of information from chronology. In this function, $f_{conv}$, the output $Y_1$ retains the same shape as the input.
Secondly, there may be relevance in the variation patterns of different components of the feature vector along the time axis, and this kind of relevance may imply correlations in different feature attributes. Based on this, we perform self-attention of the variation sequences of vector components on the feature map, ${Y_1}=(\bm{y^1_1},\dotsc, \bm{y^1_n})\in\mathbb{R}^{d \times n}$. Hence,
\begin{align}
  Q & = W_Q{Y_1}^T                                             \\
  K & = W_K{Y_1}^T                                             \\
  V & = W_V{Y_1}^T                                             \\
  H & = V \cdot \operatorname{softmax}(\frac{K^TQ}{\sqrt{d}}),
\end{align}
where $W_Q, W_k, W_V\in \mathbb{R}^{\frac{n}{2}\times d}$. Because chronologically adjacent vectors are similar, the variation sequences of vector components should have redundancy. Thus, a dimensionality reduction method similar to that used in non-local~\cite{wang2018non} is used to avoid over-fitting and reduce the amount of calculation. Finally, we apply the linear mapping $W_O\in \mathbb{R}^{n \times \frac{n}{2}}$ to $H\in \mathbb{R}^{\frac{n}{2}\times d}$ to transform it into the original vector dimension. Denoting the results of this function, $f_{atten}$, as $Y_2 =(W_OH)^T$, $Y_2$ still keeps the same shape as the input.

Finally, we leverage the contextual vector $\bm{c}$ to perform hybrid attention combining external attention and self-attention. We denote that $Y^\prime_2 =(\bm{{y^2_1}^\prime},\dotsc, \bm{{y^2_n}^\prime)}$, where $\bm{{y^2_i}^\prime} = \begin{bsmallmatrix}\bm{y^2_i}\\ \bm{c}\end{bsmallmatrix}, i\in[1, n)$. The blended attention procedure can be formulated as follows:
\begin{align}
  \hat{Q} & =\hat{W_Q}Y^\prime_2                                                                                            \\
  \hat{K} & =\hat{W_K}Y_2                                                                                                   \\
  \hat{V} & =\hat{W_V}Y^\prime_2                                                                                            \\
  A       & = \operatorname{softmax}(\frac{{\hat{K}}^T\hat{Q}}{\sqrt{d}}),\hspace{2em}A =(a_{ij})\in\mathbb{R}^{d\times d},
\end{align}
where $\hat{W_Q}, \hat{W_V}\in \mathbb{R}^{d\times 2d}, \hat{W_K}\in\mathbb{R}^{d\times d}$. The reason why $Y_2$ and $Y^\prime_2$ are both used for calculation is that all ${y^2_i}^\prime$ in $Y^\prime_2$ are spliced with the same contextual vector $\bm{c}$. If $\hat {W_Q}, \hat{W_K}$ are both obtained from the projecting of $Y^\prime_2$, then redundant calculations will be introduced in the process of begging the weight matrix $A$. For this function, $f^\prime_{atten}$, the final output is flexible, i.e,
\begin{align}
  \shortintertext{if not aggregated,}
  f^\prime_{atten}(Y_2, \bm{c}) & = \hat{V} \cdot softmax(A),
  \shortintertext{if aggregated,}
  f^\prime_{atten}(Y_2, \bm{c}) & = \hat{V} \cdot \operatorname{softmax}((\sum^d_{i = 1}a_{i1}, \dotsc, \sum^d_{i = 1}a_{id})).
\end{align}

When aggregation is required, often in consideration of the amount of calculation of concatenating multiple modules or the requirement of the shape of the final output, the output is $Y\in \mathbb{R}^{d}$; otherwise, the output belongs to $\mathbb{R} ^{d\times n}$ by default, which retains the same shape as the input of the module.

After adding the normalization operations and residual delivering, the structure of this universal network module can be expressed as,
\begin{align}
  \tilde{Y_1} & = f_{conv}(\Phi(X)+X)                              \\
  \tilde{Y_2} & = f_{atten}(\Phi(\tilde{Y_1})+\tilde{Y_1})         \\
  \tilde{Y}   & = f_{atten}^\prime(\Phi(\tilde{Y_2})+\tilde{Y_2}),
\end{align}
among them, $\Phi$ represents the joint operation of performing LayerNorm first and then using the activation function.
\subsection{Modular Network Architecture}
\label{ssb:1}
\begin{figure*}[t]
  \centering
  \includegraphics[width=0.8\linewidth]{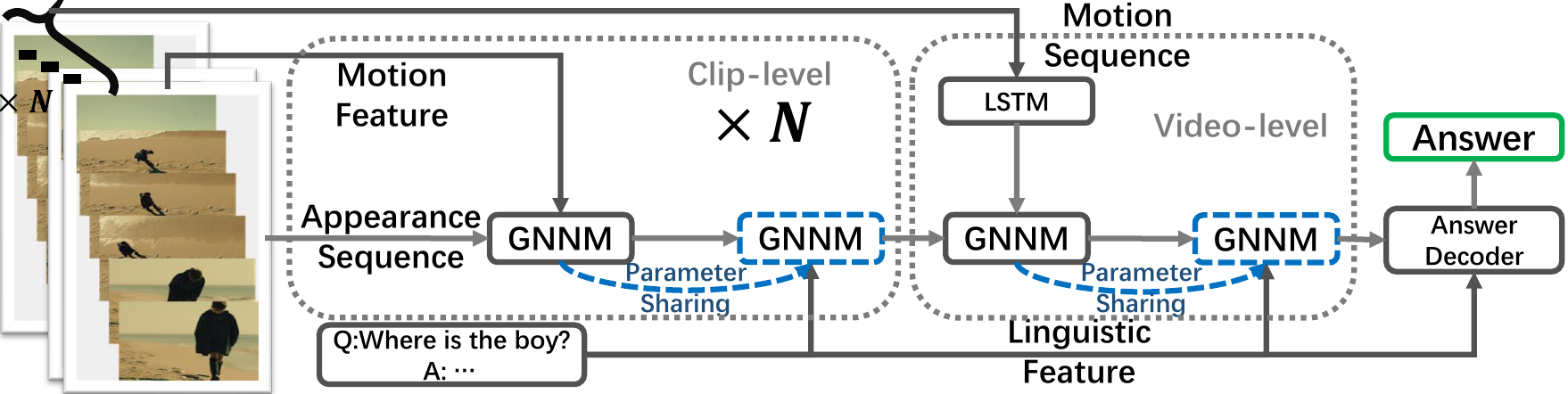}
  \caption{Our modular neural network structure that consists of clip-level and video-level organizations, and each of them contains of two generic neural modules.}
  \label{fig:2}
\end{figure*}
Inspired by HCRN~\cite{le2020hierarchical}, for video question answering tasks, we divide the overall network structure into two levels: clip-level and video-level. As shown in Fig.~\ref{fig:2}, each level is composed of two universal network modules in series. The feature embedding sequence of the video frames, ${X}=(\bm{x_1},\dotsc, \bm{x_n})\in\mathbb{R }^{d \times n}$, is taken as the input array of the first unit, and this repeats several times to process all clips of a complete video to get the sequence of aggregated clip embeddings to feed into the modular of the next level; as for the contextual vector, $\bm{c}\in\mathbb{R}^d$, we use the dynamic visual information of the video clip or video, and the natural language information of the video question for different modules respectively. In particular, for the task of the multiple-choice question, two additional modules with candidate answers served as contextual vectors have been added, containing a total of six layers.

We set the last layer of the two levels to perform the aggregation operation. The aggregation at the last layer of the video level is obviously to complete the prediction of the final result; while at the video segment level, it is mainly for reducing the computational complexity. Since the video is divided into multiple segments, the base input at the lowest unit of structure of the video level should be the sequence that all outputs of the clips combined into, viz., it will significantly increase the computing consumption without aggregating the outcomes of the video segment level.

For different generic network modules at the same level, the fundamental attribute that distinguishes each other lies in their input and their position in the network; while the parameter quantity, the scale of the network, is constant and only predicated on the shape of the feature map. Therefore, from the implementation point of view, it is feasible to reuse a universal network module through the parameter sharing along with changing its input to enable it to be a logically distinct module.

\subsection{Guess Before You Ask: Two Stage Training}
\label{sec:two_stage}
In our case, we find our model tends to be overly dependent on the linguistic features, partly because of the datasets whose questions and their corresponding answers are highly relevant and on which the model can predict the answers with little visual information; partly because the ultimate purpose of the model is to match the correct answers which are encoded as linguistic embeddings, it is easier to take advantage of the linguistic features instead of visual features since features in the same modality are more relevant.

Therefore, we divide the training process into two stages for further ablation study. In the first stage, we do warm-up training without linguistic conditional features and the linguistic feature of the question in the answer decoder. We aim to get a candidate set of a certain video; this course is, to some extent, a procedure like an object detection. After that, in the second stage, we utilize fully integrated features to fine-tune the model while the learning rate of the parameters of the two generic neural network modules with the visual feature is set to a small value.
\subsection{Answer Decoder and Loss Functions}
Following~\cite{le2020hierarchical,park2021bridge}, we use three answer decoders for three types of questions in our adopted datasets.
For open-ended questions, we treat them as multi-label classification problems and use the cross-entropy as its loss function. Given the final output of the model, $\tilde{Y}$, and the question embedding $q$, the answering decoder for question set with the corresponding answer space $\mathcal{A}$ can be expressed as,
\begin{align}
  \bm{y}          & =\sigma\left(W_{\tilde{Y}}\left[\tilde{Y}; W_{q} \bm{q}+\bm{b}\right]+\bm{b}\right) \\
  \bm{y^{\prime}} & =\sigma\left(W_{y} \bm{y}+\bm{b}\right)\label{eq:open1}                             \\
  \bm{p}          & =\operatorname{softmax}\left(W_{y^{\prime}} \bm{y^{\prime}}+\bm{b}\right),
\end{align}
where $\sigma$ stands for the activate function and $\bm{p} \in \mathbb{R}^{|\mathcal{A}|}$ represents the label probabilities. Particularly, while warm-up training, $\bm{p^\prime}$ will be calculated as follows:
\begin{align}
  \bm{\tilde{y}} & =\sigma\left(W_{\tilde{Y}} \tilde{Y}+\bm{b}\right)\label{eq:open2}       \\
  \bm{p^\prime}  & =\operatorname{softmax}\left(W_{\tilde{y}} \bm{\tilde{y}}+\bm{b}\right).
\end{align}
For the repetition action count task in TGIF-QA, we take it as a linear regression question and adopt the Mean Squared Error (MSE) as its loss function. Taking $\bm{y^{\prime}}$ in~(\ref{eq:open1}) and $\bm{\tilde{y}}$ in~(\ref{eq:open2}) as inputs correspondingly, the results can be caluculated as follows:
\begin{align}
  o          & =\operatorname{round}\left(W_{y^{\prime}} \bm{y^{\prime}}+\bm{b}\right) \\
  {o^\prime} & =\operatorname{round}\left(W_{\tilde{y}} \bm{\tilde{y}}+\bm{b}\right).
\end{align}
For the multiple choice question, there is extra linguistic information, answer candidates $a$. We treat the task a as multi-label classification problem as same as the open-ended question task but with hinge loss, $\operatorname{max}(0, 1 + p^i - p^{GT})$, adopted to compute the loss between the ground truth answer, $p^{GT}$, and the other candidates, $p^i$. The answer decoder can be formulated as,
\begin{align}
  \bm{y}          & =\left[W_{\tilde{Y}}\tilde{Y}+\bm{b}; W_{q} \bm{q}+\bm{b};W_{a} \bm{a}+\bm{b}\right] \\
  \bm{y^{\prime}} & =\sigma\left(W_{y} \bm{y}+\bm{b}\right)                                              \\
  \bm{p_a}        & =W_{y^{\prime}} \bm{y^{\prime}}+\bm{b}.
\end{align}
Similarly, while warm-up training, $\bm{p^\prime}$ will be calculated as follows:
\begin{align}
  \bm{y}          & =\left[W_{\tilde{Y}}\tilde{Y}+\bm{b}; W_{a} \bm{a}+\bm{b}\right] \\
  \bm{\tilde{y}}  & =\sigma\left(W_{y} \bm{y}+\bm{b}\right)                          \\
  \bm{p_a^\prime} & =W_{\tilde{y}} \bm{\tilde{y}}+\bm{b}.
\end{align}
\section{Experiment and Evaluation}
\subsection{Datasets}
We conduct experiments on three popular video QA datasets: TGIF-QA~\cite{jang2019video}, MSVD-QA~\cite{xu2017video}, and MSRVTT-QA~\cite{xu2017video}.
\subsubsection{TGIF-QA}
The dataset contains $165\,000$ QA pairs for the animated GIFs from TGIF dataset~\cite{li2016tgif}. The questions can be divided into four categories: (a) counting the number of a repetitive action, (b) identifying a repetitive action, (c) identifying the state before (or after) another state, and (d) frame question and answering that can be answered from one of frames; among them, (a) and (d) are open-ended questions, and (b) and (c) are multiple-choice questions.
\subsubsection{MSVD-QA}
The dataset consists of a total number of $1\,970$ video clips and $50\,505$ question-answer pairs, and the average video length is about ten seconds. The dataset includes five question types: What, Who, How, When, and Where. The proportions of videos in the training, testing, and validation splits are $61\%, 13\%$, and $26\%$, respectively.

\subsubsection{MSRVTT-QA}
The dataset holds $10\,000$ videos and $243\,000$ question-answer pairs. The classes of questions are the same as MSVD-QA dataset, but the contents of the videos in MSRVTT-QA are more complicated and the lengths of the videos are longer from $10$ to $30$ seconds. The proportions of videos in the training, testing, and validation splits are $65\%, 30\%$, and $5\%$, respectively.

ccccc
\subsection{Complexity Analysis}
\label{sec:complexity}
For real-world neural networks, memory accesses play a more important role than the number of computations since a single memory access is much slower than a single computation. Besides, time complexity has little value in the assessment of the speed the models run, considering the distant parallel processing capabilities of GPUs. Therefore, the generally adopted measure for the complexity of a model is the number of its parameters. The lower bound (preserving the partial derivative intermediate terms during the back-propagation process will occupy more space.) of the overall space complexity of our module can be expressed as,
\begin{align}
  Space \sim \Omega(14d^2+12d+7),\label{eq:1}
\end{align}
generally, since the convolutional layer and the LayerNorm layer have a much smaller impact on the space complexity than the full connection, with only the highest-order terms retained, (\ref{eq:1}) can also be expressed as,
\begin{align}
  Space \sim \Omega(14d^2).
\end{align}
For the overall network, as described in the Section~\ref{ssb:1}, according to whether there is candidate answer information, its structure can be of six layer or four layer; the parameter sharing is carried out in the two levels separately, and the number of universal network modules can be reduced to two, for segment level and video level correspondingly, which significantly reduces the occupation of computing space.
\subsubsection{Space Complexity}
The calculation of the space complexity of the feedforward neural network includes two parts, gradient descent and back-propagation. Suppose that the number of parameters (scalars) of a neural network is $m$, then its space complexity is $\Omega(2m+1)$, which involves all the parameters of the neural network, the corresponding gradients of the trainable parameters, and the learning rate. In a neural network, most of the parameters are trainable, so we approximate that the gradient data is equal to the trainable parameters. It is worth noting that the space complexity of back-propagation depends on the specific implementation of solving partial derivatives; here $2m+1$ is only the lower bound, if you need to store the intermediate terms of partial derivatives, it will occupy more space.

We use the number of parameters to measure the space complexity of the module network; in fact, the GPU uses the form of $float32$ to store parameters, so each parameter corresponds to $32~bits$. Here we will analyze the space complexity of our universal network module according to its three-tier structure. First, in the convolutional layer,
\begin{align}
  Params_{conv} = C_{in} \times S \times C_{out} + C_{out},
\end{align}
among them, $C_{in}$ is the number of channels in the input feature map; $C_{out}$ is the number of channels in the output feature map; $S$ is the parameter amount of the convolution kernel, for the $3\times 1$ convolution kernel, its parameter amount is three; the last item, $C_{out}$ , represents the number of bias parameters. In our case, the input and output channels are both one, and the bias is not used; therefore, $Params_{conv} = 3$. Second, in the feature dimension attention procedure, there are four fully connected layers; the parameter quantity of each fully connected layer can be expressed as,
\begin{align}
  Params_{linear}= L_{in}*L_{out}+L_{out},
\end{align}
where $L_{in}$ and $L_{out}$ represent the feature dimensions of input and output respectively; the last item, $L_{out}$, represents the bias parameters, which are not used here. Therefore, according to the specific settings described in the method section of the four fully connected layers, its parameter amount can be obtained as,
\begin{align}
  Params_{atten} = 4 \times (d \times \frac{d}{2}).
\end{align}
Finally, for the blended attention part, there are a total of three fully connected layers. According to the method description, its parameter amount can be calculated as,
\begin{align}
  Params_{atten}^\prime = d\times d+2 \times (d \times 2d).
\end{align}
In addition, considering the scaling parameters and biases in the three LayerNorm layers of the module, there are a total of $3\times 2d$ parameters, then the amount of parameters of the whole module can be expressed as,
\begin{align}
  Params = 7d^2+6d+3,
\end{align}
where the non-trainable parameters in the activation function and the hyperparameters in the specific implementation of the algorithm are ignored.

Therefore, the lower bound (preserving the partial derivative intermediate terms during the back-propagation process will occupy more space.) of the overall space complexity of the module can be expressed as,
\begin{align}
  Space \sim \Omega(14d^2+12d+7),\label{eq:1}
\end{align}
Generally, since the convolutional layer and the LayerNorm layer have a much smaller impact on the space complexity than the full connection, with only the highest-order terms retained, (\ref{eq:1}) can also be expressed as,
\begin{align}
  Space \sim \Omega(14d^2).
\end{align}

For the overall network, as described in the Section~\ref{ssb:1}, according to whether there is candidate answer information, its structure can be of six layer or four layer; the parameter sharing is carried out in the two levels separately, and the number of universal network modules can be reduced to two, for segment level and video level correspondingly, which significantly reduces the occupation of computing space.

\subsection{Implementation Details}
Following the previous work~\cite{le2020hierarchical,park2021bridge}, we divide the video $\mathcal{V}$ of $L$ frames into $N$ uniform length clips and use pre-trained ResNet, ResNeXt, and GloVe to extract appearance, motion, lingustic features respectively.
For all the experiments, we use the official split of training, validation, and testing sets of each dataset. We divide the video into eight clips containing $16$ frames in each clip by default. We train our model for $25$ epochs with a batch size of $128$ for TGIF-QA~\cite{jang2019video} and MSVD-QA~\cite{xu2017video} datasets, and of $64$ for MSRVTT-QA~\cite{xu2017video} dataset. The learning rate is set to $10^{-4}$ and decayed by half for every five epochs. The reported results are at the epoch showing the best validation accuracy. For vector dimension of generic neural network module, we set it to $512$ by default.


\subsection{Experimental Results}
We compare our proposed model with several state-of-the-art methods on TGIF-QA, MSVD-QA, and MSRVTT-QA datasets. On the TGIF-QA dataset, we compare our model with~\cite{jang2017tgif,gao2018motion,li2019beyond,fan2019heterogeneous,huang2020location,jiang2020divide,le2020hierarchical,lei2021less}. The results over four tasks are presented in Tab.~\ref{tab:my-table1}, including repeating action, state transition, frameQA, and repetition counting. The results show that our model achieves state-of-the-art performance and significantly outperforms the existing methods on repeating action and state transition tasks.

\begin{table}[ht]
  \begin{center}
    \caption{Performance comparison on TGIF-QA dataset. The lower the better for count.}
    \label{tab:my-table1}
    \resizebox{0.45\textwidth}{!}{
      \begin{tabular}{lcccc}
        \hline
        \textbf{Model}                     & \textbf{Action} & \textbf{Transition} & \textbf{Frame} & \textbf{Count} \\ \hline
        ST-TP~\cite{jang2017tgif}          & 62.9            & 69.4                & 49.5           & 4.32           \\
        Co-Mem~\cite{gao2018motion}        & 68.2            & 74.3                & 51.5           & 4.10           \\
        PSAC~\cite{li2019beyond}           & 70.4            & 76.9                & 55.7           & 4.27           \\
        HME~\cite{fan2019heterogeneous}    & 73.9            & 77.8                & 53.8           & 4.02           \\
        L-GCN~\cite{huang2020location}     & 74.3            & 81.1                & 56.3           & 3.95           \\
        QueST~\cite{jiang2020divide}       & 75.9            & 81.0                & 59.7           & 4.19           \\
        HCRN~\cite{le2020hierarchical}     & 75.0            & 81.4                & 55.9           & 3.82           \\
        ClipBERT~\cite{lei2021less}         & 82.8            & 87.8                & 60.3           & -              \\
        Bridge2Answer~\cite{park2021bridge} & 75.9            & 82.6                & 57.5           & 3.71           \\
        \hline
        \textbf{Ours}                      & 90.8            & 93.9                & 56.7           & 3.81           \\ \hline
      \end{tabular}}
  \end{center}
\end{table}

For MSVD-QA and MSRVTT-QA datasets, Tab.~\ref{tab:my-table2} shows the comparisons between our model against the state-of-the-art methods~\cite{xu2017video,chowdhury2018hierarchical,gao2018motion,fan2019heterogeneous,huang2020location,jiang2020divide,le2020hierarchical,park2021bridge}. Our model achieves $38.3\%$ and $36.1\%$ accuracy. It is clear that our proposed method consistently outperforms all SOTA models over the MSVD-QA task. The lower accuracy on MSRVTT-QA is mostly due to the long videos in MSRVTT-QA being still divided into eight clips in our experiment setting to ensure consistency.
\begin{table}[ht]
  \centering
  \caption{Performance comparison for open-ended questions on
    MSVD-QA and MSRVTT-QA datasets.
  }
  \label{tab:my-table2}
  \begin{tabular}{lcc}
    \hline
    \textbf{Model}                       & \textbf{MSVD-QA} & \textbf{MSRVTT-QA} \\ \hline
    AMU~\cite{xu2017video}               & 32.0             & 32.5               \\
    HRA~\cite{chowdhury2018hierarchical} & 34.4             & 35.0               \\
    Co-Mem~\cite{gao2018motion}          & 31.7             & 31.9               \\
    HME~\cite{fan2019heterogeneous}      & 33.7             & 33.0               \\
    L-GCN~\cite{huang2020location}       & 34.3             & -                  \\
    QueST~\cite{jiang2020divide}         & 36.1             & 34.6               \\
    HCRN~\cite{le2020hierarchical}       & 36.1             & 35.6               \\
    Bridge2Answer~\cite{park2021bridge}   & 37.2             & 36.9
    \\ \hline
    \textbf{Ours}                        & \textbf{38.3}    & \textbf{36.1}      \\ \hline
  \end{tabular}
\end{table}

\subsection{Ablation Study}
\begin{table}[ht]
  \centering
  \caption{Ablation studies results on MSVD-QA, MSRVTT-QA and TGIF-QA dataset. Default model config. (*): six layers of GNNM for multiple-choice questions and four layers for others.}
  \resizebox{0.45\textwidth}{!}{
    \label{tab:my-tablesta}
    \begin{tabular}{@{\extracolsep{\fill}}lcccccc}
      \hline
      \textbf{Model}                  & \textbf{MSV.}  & \textbf{MSR.}  & \textbf{Act.} & \textbf{Tran.} & \textbf{Frm.}  & \textbf{Cnt.} \\ \hline
      \textbf{Two-stage training}     &                &                &               &                &                &               \\
      \hspace{3mm}Warm-up training    &                &                &               &                &                &               \\
      \hspace{6mm}w/o questions       & 18.61          & 13.14          & 89.09         & 92.24          & 34.20          & 4.45          \\
      \hspace{3mm}Fine-tune training  &                &                &               &                &                &               \\
      \hspace{6mm}w/ questions        & \textbf{38.34} & \textbf{36.08} & 90.58         & 93.17          & \textbf{56.74} & \textbf{3.81} \\ \hline
      \textbf{Regular training}       &                &                &               &                &                &               \\
      \hspace{3mm}Default config. (*) & 37.57          & 35.76          & 90.11         & 93.71          & 56.05          & 3.86          \\ \hline
      \textbf{Parameter sharing}      &                &                &               &                &                &               \\
      \hspace{3mm}Two modulars        & \textbf{38.10} & \textbf{36.13} & 90.77         & 93.14          & \textbf{56.63} & 3.86          \\ \hline
      \textbf{Vector dimension}       &                &                &               &                &                &               \\
      \hspace{3mm}Two-stage training  &                &                &               &                &                &               \\
      \hspace{6mm}$d=256$             & 37.90          & 35.93          & 90.27         & 93.04          & 56.03          & 3.82          \\
      \hspace{6mm}$d=128$             & 37.55          & 35.77          & 90.17         & 93.34          & 55.28          & 3.84          \\
      \hspace{3mm}Regular training    &                &                &               &                &                &               \\
      \hspace{6mm}$d=256$             & 36.83          & 35.23          & 90.45         & 93.14          & 55.43          & 3.84          \\
      \hspace{6mm}$d=128$             & 35.36          & 34.31          & 90.63         & 93.94          & 53.29          & 3.89          \\
      \hspace{6mm}$\underline{d=0}$   & -              & -              & 72.67         & 74.45          & -              & -             \\ \hline
    \end{tabular}}
\end{table}
To explore the potential of our network and contribute to a deeper understanding of our generic neural network module, ablation experiments are performed. The details are shown in Tab.~\ref{tab:my-tablesta}.
\paragraph{$d=256$ or $128$}
Generally, the compatibility of the number of model parameters and the scale of input information decides whether the model will fit the data; overly complex models typically have low bias and high variance~\cite{lever2016points}. Therefore, we conduct an ablation study on the vector dimension to see if our parameter amount is suitable. As a result, for both vector dimensions of 256 and 128, there is only a slight decline in the accuracy, especially for action and transition tasks.
\paragraph{$d=0$}
To explore the reasons further, we have an extreme experiment setting the dimension to zero, which means there are only conditional features that are fed into the answer decoder to solve the open-ended questions.

As mentioned in Section~\ref{sec:two_stage}, we study the effectiveness of two-stage training with intention of promoting our model to learn both linguistic and visual features.
\paragraph{w/o questions}
As is specified in Section~\ref{sec:two_stage}, in warm-up training, the question embeddings will not be fed into the model.
\paragraph{w/ questions}
In the fine-tune training process, the learning rates of parameters of clip-level motion condition and video-level condition modules are set to $0.05$ times compared to the default configuration. And the all learning rates decay is set to $0.5$ every three epochs. The experiment results prove that our two-stage training indeed works. This kind of method is similar to some pre-trained models but requires no additional dataset. And, as such, it is unable to learn more general representations due to the limitation of the dataset. Based on the results of two-stage training and ``$\underline{\bm{d=0}}$'', we can infer that the ground truth and their corresponding candidate answers are highly relevant, and the model can predict the answers with bare linguistic information.
It is to say that the results of these two tasks, action and transition, can not justify the effectiveness of our model strongly in embedding fusing.

\paragraph{two modulars}
When applying parameter sharing, the modules in the same level (e.g., clip-level or video-level) will invoke a physically identical model; to some extent, the model here has a strong resemblance to the kernel of the convolutional neural network. Beyond our expectation, the two-modular model stays at the same level as the default model and outperforms it on several tasks.

\section{Conclusion}
We present a new generic neural modular network aiming at finding a feasible and efficient approach to Video QA tasks. Our generic module fuses multimodal features conditioned on chronological clues and blended attention mechanisms. By easily connecting the sequence-to-sequence modules in series or parallel, we can form intricate models with little effort. Besides, we validate the effectiveness of the parameter sharing and two-stage training strategies, and both of them show promising performance. Given constrained computational resources and a growing multimodal task scale, our generic neural modular network presents ongoing competitiveness; it provides a glimpse into the sought-after end of simple and efficient modular neural networks.

\bibliographystyle{IEEEabbre}
\bibliography{icme2022template}

\end{document}